\ifdefined\pdfminorversion\pdfminorversion=7\fi
\documentclass[conference]{IEEEtran}
\IEEEoverridecommandlockouts

\usepackage{times}
\usepackage{cite}
\usepackage{amsmath,amssymb,amsfonts}
\usepackage{graphicx}
\usepackage{booktabs}
\usepackage{textcomp}
\usepackage{xcolor}
\definecolor{projpink}{HTML}{D81B60}
\usepackage[hidelinks,bookmarks=true,pdftitle={Navigation Code World Model: Decoupling Semantic Path Planning from Visual Rendering}]{hyperref}

\renewcommand{\footnoterule}{\kern -3pt \hrule width 0.4\columnwidth \kern 2.6pt}
\renewcommand{\IEEEbibitemsep}{4.5pt plus 1.5pt minus 1pt}
\makeatletter
\long\def\@makefntext#1{\noindent #1}
\newcommand{\blfootnote}[1]{%
  \begingroup
  \renewcommand\thefootnote{}%
  \footnote{#1}%
  \addtocounter{footnote}{-1}%
  \endgroup
}
\let\oldthebibliography\thebibliography
\renewcommand{\thebibliography}[1]{\oldthebibliography{#1}\fontsize{9pt}{11.2pt}\selectfont}
\makeatother

\begin{document}

\title{Navigation Code World Model: Decoupling Semantic Path Planning from Visual Rendering\vspace{0.35em}}

\author{
\IEEEauthorblockN{
Hongjin Chen$^{1,2,*}$,
Liangwei Qin$^{1,*}$,
Weicheng Feng$^{3}$,
Shaojie Ren$^{2,4}$,
Jingwen Sun$^{2,5}$,
Zezhi Liu$^{2,6}$, \\[0.3em]
Liang Xu$^{2}$,
Tonghua Su$^{1}$,
Chen Gao$^{2,7,\dagger}$,
Yong Li$^{7}$,
Zhibo Chen$^{2,5,\dagger}$
}
\vspace{0.35em}
\IEEEauthorblockA{
$^{1}$Harbin Institute of Technology \hspace{0.9em}
$^{2}$Zhongguancun Academy \hspace{0.9em}
$^{3}$Shandong University \\[0.18em]
$^{4}$University of the Chinese Academy of Sciences \hspace{0.9em}
$^{5}$University of Science and Technology of China \\[0.18em]
$^{6}$Nankai University \hspace{0.9em}
$^{7}$Tsinghua University
}
}

\makeatletter
\twocolumn[{%
\renewcommand\twocolumn[1][]{#1}%
\maketitle
\vspace{-1.75em}
\begin{center}
\centering
\includegraphics[width=0.815\textwidth]{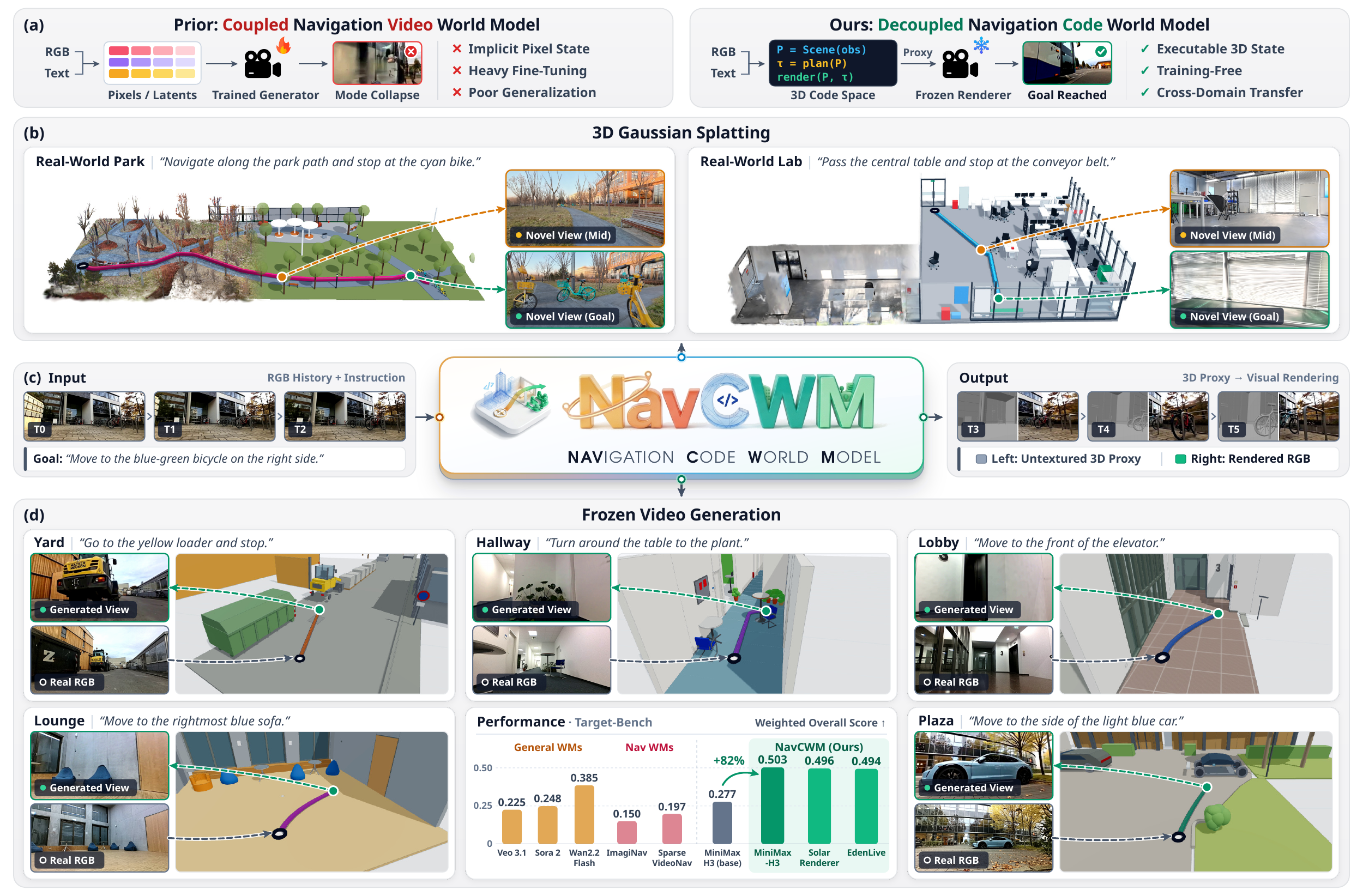}
\par\vspace{-0.55em}
{\def\@captype{figure}%
\caption{\textbf{Overview of NavCWM.} \textbf{(a)}~Unlike coupled video world models, our decoupled code-world-model paradigm separates executable 3D code-space planning from a frozen renderer, training-free. \textbf{(b)}~Lifting real-world 3DGS scans (\emph{Park}, \emph{Lab}; split 3DGS/code) into executable 3D scene code enables long-horizon navigation path planning and mid/goal novel views. \textbf{(c)}~Causal RGB (T0--T2) and goal text yield untextured 3D proxies guiding rendered RGB rollouts (T3--T5). \textbf{(d)}~Five indoor/outdoor frozen-video rollouts (\emph{Yard}, \emph{Hallway}, \emph{Lobby}, \emph{Lounge}, \emph{Plaza}; \emph{Real RGB} $\to$ 3D path $\to$ \emph{Generated View}, overviews colored for clarity) and Target-Bench scores lifting MiniMax-H3 from 0.277 to 0.503 WO (+82\%; 0.494--0.503 across three renderers).}%
\label{fig:pipeline}}
\vspace{-0.60em}
\end{center}%
}]
\makeatother
\thispagestyle{plain}

\begin{abstract}
Embodied\blfootnote{$^{*}$Equal contribution. $^{\dagger}$Corresponding authors.} navigation requires grounded spatial reasoning to reach semantic goals,
motivating world models that simulate future observations for planning. Yet
current video world models often couple spatial state with visual appearance in
pixel or latent rollouts, where mode collapse and limited cross-domain
generalization remain persistent challenges. We introduce Navigation Code World
Model (NavCWM), a training-free framework that decouples semantic path planning
from visual rendering through executable 3D scene code. Given causal
egocentric observations and a language instruction, its front end
synthesizes, visually refines, and verifies a metric scene program against observed views before
planning a collision-free path toward the instructed target. This front end
drives two interchangeable back ends: frozen video generators steered by
untextured 3D proxies and 3D Gaussian Splatting (3DGS) for direct novel-view
synthesis. On Target-Bench, NavCWM lifts frozen MiniMax-H3 from 0.277 to 0.503
in weighted overall (WO) score, leading all zero-shot baselines and
surpassing the fine-tuned reference (0.478 WO). The same scene code and proxies
transfer zero-shot to SolarRenderer (0.496 WO) and EdenLive (0.494 WO). Proxy ablations trace
these gains to complete untextured 3D geometry rather than sparse landmarks,
single-plane motion, or colored meshes. In real-world indoor and outdoor 3DGS
scans, executable spatial memory outperforms episodic memory in long-horizon
arrival and enables instruction-driven data generation. To our knowledge, NavCWM is the first code world model for
navigation, pointing toward interpretable, verifiable, and renderer-agnostic
world representations for mobile robots. Project page: \href{https://chenhongjin811.github.io/NavCWM/}{\textcolor{projpink}{\texttt{https://chenhongjin811.github.io/NavCWM/}}}.
\end{abstract}

\vspace{-0.3em}

\section{Introduction}
\label{sec:intro}
Guiding a mobile robot to language-specified goals relies on grounded spatial
understanding and predictive
foresight~\cite{anderson2018vln,shah2022lmnav,zhang2024navid}. To provide such
foresight, embodied world models~\cite{hou2026wmsurvey} have progressed from
compact latent dynamics and abstract transition
predictors~\cite{ha2018worldmodels,lecun2022path,hafner2025dreamerv3} to
large-scale video world simulators~\cite{yang2024unisim,sora2024,agarwal2025cosmos}.
Building on this paradigm, recent navigation world models forecast future
egocentric observations or video-latent transitions to guide planning and
action~\cite{nwm,imaginav,sparsevideonav,worldvln}. However, video-based world
models that predict scene geometry and visual appearance jointly in pixel or
latent space still exhibit rollout drift and mode collapse~\cite{nwm,sparsevideonav}, limited cross-domain transfer~\cite{nwm},
and fragile zero-shot metric planning~\cite{targetbench} even over short
horizons. Over longer horizons, the absence of an explicit global 3D and
semantic state that supports collision checking makes it difficult to ground
language instructions in scene-wide geometry, maintain collision safety without
drift, and scale to large environments.

We rethink navigation world models through robot modeling and simulation,
viewing them as internal simulators~\cite{hou2026wmsurvey} and separating an
explicit 3D scene and motion model from visual sensor rendering, echoing modular
mapping and planning~\cite{elfes1989occupancy,hart1968astar,gervet2023navigating}.
Realizing this separation in real environments exposes a recurring gap across
fragmented 3D representations~\cite{deng2025best}:
hand-authored simulators~\cite{kolve2017ai2thor,li2023behavior1k} support
explicit geometry and planning but rely on synthetic assets and face a
sim-to-real gap~\cite{zhao2020simtoreal}. Conversely, 3D Gaussian Splatting
(3DGS) reconstructions~\cite{kerbl20233dgs} preserve scene appearance yet lack
explicit collision geometry and traversability bounds for planning, whereas language-queryable 3D semantic maps~\cite{huang2023vlmaps},
scene graphs~\cite{gu2024conceptgraphs,werby2024hovsg}, and episodic memory~\cite{anwar2025remembr}
lack visual foresight. Meanwhile, video world models offer
strong visual generation without an explicit global scene model. Each approach
addresses only part of the problem: what is missing is an observation-grounded
real-to-sim scene modeling layer that explicitly encodes 3D geometry, collision
bounds, and open-vocabulary semantics while remaining directly executable to
drive decoupled visual renderers.

In fact, robotics simulators already define scenes programmatically, though
manual assets scale poorly. Recent advances in structured and agentic 3D
modeling~\cite{hu2024scenecraft,avetisyan2024scenescript,zhang2025scenelanguage},
however, now make \emph{executable 3D scene code} a practical path. Operating in \emph{3D code space} offers three properties: first,
\emph{structural abstraction}, distilling partial views or scans into clean,
appearance-free visual, collision, and semantic primitives rather than implicit
latents or raw unstructured geometry; second, \emph{programmatic editability},
exposing named parameters that agents can iteratively inspect and patch against
observed views, unlike neural weights or dense point clouds; and third,
\emph{executable modularity}, uniting symbolic querying, planning, and proxy
synthesis in one program to drive interchangeable renderers. While recent code
world models span video, proxy rendering, physics, and
memory~\cite{codevideomodel,pwm2026,edenlive,solarrenderer,cwm,codeasworlds,agentgarten,eidetic2026},
grounding verified 3D code in real egocentric observations for semantic navigation
remains open.

We address this gap with \textbf{Navigation Code World Model} (\textbf{NavCWM}),
a training-free framework shown in Fig.~\ref{fig:pipeline} and detailed in
Sec.~\ref{sec:method} that separates a code \emph{front end} for scene modeling
and semantic path planning from an interchangeable visual \emph{back end}. A coding agent translates causal
egocentric observations and a language instruction into an executable 3D scene
program capturing layout, collision bounds, and target semantics before
refining and verifying it via render-and-compare against observed views. NavCWM then plans a
collision-free semantic path in code space and produces an untextured white-box
proxy video that steers a frozen video generator alongside the
current frame and instruction for photorealistic visual foresight. We also lift
captured real-world 3DGS scans once into executable 3D spatial memory,
supporting long-horizon vision-language navigation (VLN) planning and 3DGS
novel-view synthesis for instruction-driven data generation.

Our contributions are threefold:
\begin{enumerate}
\item \textbf{A new code-space paradigm for navigation world models.}
NavCWM represents environment geometry, collision bounds, and target semantics as verified executable 3D
scene code refined via render-and-compare. This front end separates
collision-free path planning from interchangeable visual back ends spanning
frozen video generators and 3DGS.
\item \textbf{Decoupled proxy-conditioned architecture and zero-shot generalization.}
From causal egocentric observations, untextured 3D proxies lift frozen
MiniMax-H3 from 0.277 to 0.503 in weighted overall (WO) score via early-step steering and transfer
zero-shot to SolarRenderer~\cite{solarrenderer} at 0.496 and EdenLive~\cite{edenlive} at 0.494, surpassing all
zero-shot baselines and exceeding fine-tuned Wan2.2-5B in WO (0.478).
Controlled ablations isolate complete untextured 3D geometry over sparse
landmarks, single-plane motion, and colored meshes as the key conditioning signal.
\item \textbf{Executable spatial memory for real-world VLN.}
Lifting a real-world 3DGS scan once into executable 3D code enables single-turn grounding that outperforms ReMEmbR~\cite{anwar2025remembr} in long-horizon metric
arrival and token efficiency, while driving 3DGS novel-view synthesis for
instruction-driven data synthesis beyond trajectory-first captioning~\cite{yu2026onescan}.
\end{enumerate}

\section{Related Work}
\label{sec:related}
\begin{figure*}[t]
  \centering
  \includegraphics[width=0.92\textwidth]{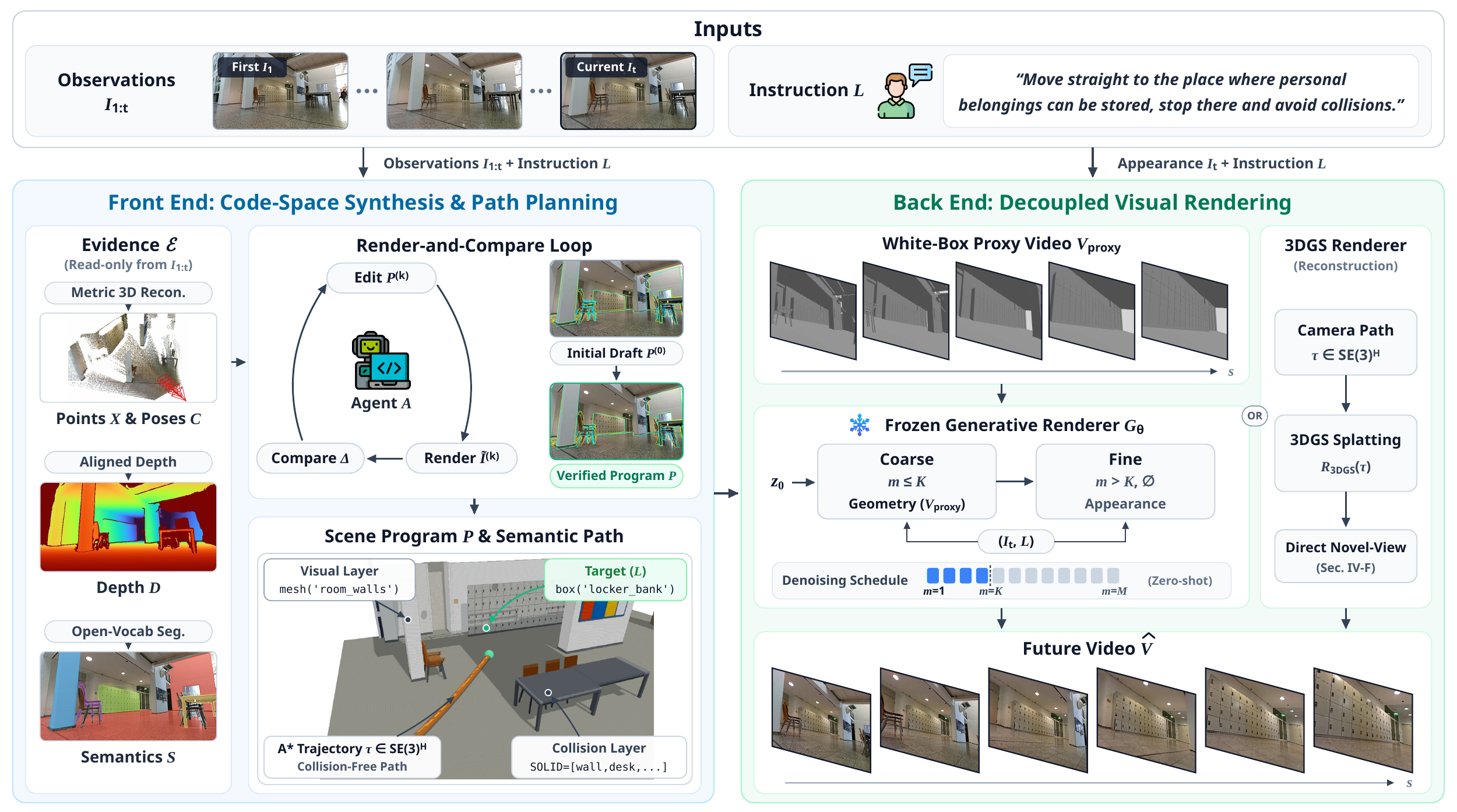}
  \vspace{-0.45em}
  \caption{\textbf{Decoupled NavCWM architecture.}
  \emph{Front End:} Given causal observations $I_{1:t}$ and instruction $L$, coding agent $\mathcal{A}$ grounds read-only multimodal evidence $\mathcal{E}$ into a verified executable 3D scene program $P$ via render-and-compare to plan a collision-free A* trajectory $\tau$.
  \emph{Back End:} Either untextured structural proxy video $V_{\text{proxy}}\!=\!\mathcal{R}(P,\tau)$ steers frozen renderer $\mathcal{G}_\theta$ alongside $(I_t, L)$ during early denoising steps $m\!\le\!K$ before switching to null proxy conditioning $\varnothing$ in late steps $m\!>\!K$, or $\tau$ directly drives 3DGS rendering $\mathcal{R}_{\text{3DGS}}(\tau)$, producing future video $\hat{V}$.}
  \label{fig:method}
  \vspace{-0.5em}
\end{figure*}

\paragraph{Navigation World Models}
Navigation world models have evolved from unvisited RGB-D panorama synthesis~\cite{koh2021pathdreamer} and abstract mental planning~\cite{wang2023dreamwalker} to high-capacity video forecasting. Action-conditioned models such as NWM~\cite{nwm} score candidate routes via egocentric video rollouts, while language-guided extensions pair visual imagination with inverse dynamics~\cite{imaginav} or predict sparse frames beyond the current view~\cite{sparsevideonav}. Related models apply test-time spatial search for vision-language reasoning~\cite{yang2025mindjourney} or couple short-horizon latent world-state prediction with closed-loop waypoint execution~\cite{worldvln}. Yet mapless evaluation on Target-Bench~\cite{targetbench} shows that encoding geometry, clearance, and texture jointly in pixel or latent space still demands task-specific training for metric accuracy across domains. NavCWM breaks this coupling by maintaining an explicit 3D state, enabling deterministic planning without fine-tuning.

\paragraph{Spatial Representations for Navigation}
Semantic mapping anchors object categories and language in metric structure, progressing from closed-set episodic maps~\cite{chaplot2020semexp} and open-vocabulary feature maps~\cite{huang2023vlmaps} to 3D scene graphs~\cite{gu2024conceptgraphs,werby2024hovsg}, vision-language frontier maps~\cite{yokoyama2024vlfm}, and retrieval-augmented memory~\cite{anwar2025remembr}, yet remains a passive memory rather than an executable visual simulator. In parallel, 3DGS reconstructions~\cite{kerbl20233dgs} preserve real-world appearance, enabling One Scan is Enough~\cite{yu2026onescan} to generate VLN trajectories on a clearance map with post-hoc captioning, but without an editable semantic program for arbitrary new instructions. NavCWM links semantic planning to photorealistic renderers via one verified program for instruction-first planning and visual synthesis.

\paragraph{Code World Models}
Executable programs serve as embodied control interfaces~\cite{codeaspolicies}, symbolic transition models~\cite{worldcoder}, and structured 3D scene representations from text or scans~\cite{hu2024scenecraft,avetisyan2024scenescript,zhang2025scenelanguage}. Extending these programs to visual simulation, Code Video Model~\cite{codevideomodel} steers frozen video generators via early-step Three.js previews, Code World Model~\cite{cwm} maintains persistent state for fine-tuned rendering, and Programmable World Model~\cite{pwm2026} compiles stateful 3D bounding boxes into video controls. Other frameworks pair game or simulator worlds with real-time renderers~\cite{zhang2026matrixgame4,agentgarten}, discover verifiable MuJoCo programs for physical reasoning~\cite{codeasworlds}, or couple geometric blockouts with 3D Gaussian appearance memory~\cite{eidetic2026}. SolarRenderer~\cite{solarrenderer}, EdenLive~\cite{edenlive}, and Code World Model~2.0~\cite{wang2026cwm2} advance proxy-conditioned rendering, while WorldDraft~\cite{hou2026worlddraft} diagnoses appearance leakage from conflicting drafts. Unlike these synthetic or draft-rendering systems, NavCWM grounds verified 3D code in causal real observations for collision-free mobile navigation and 3DGS spatial memory.

\section{Method}
\label{sec:method}

Given a causal egocentric observation history $I_{1:t}$ ending at the current
frame $I_t$ and a language navigation instruction $L$, our goal is to
synthesize a future navigation video $\hat{V}$ that advances along a physically
valid trajectory toward the semantic target. Whereas existing
language-conditioned video world models learn a joint mapping
$p(\hat{V}\mid I_{1:t}, L)$ in pixel or latent space, NavCWM decouples spatial
modeling and path planning from visual appearance synthesis by introducing an
executable scene program $P$ as the explicit world state:
\begin{equation}
\begin{aligned}
\text{\emph{Front End:}} &\;\; (I_{1:t}, L) \xrightarrow{\;\mathcal{A},\,\mathcal{E}\;} P \xrightarrow{\;\mathrm{Plan}\;} \tau, \\
\text{\emph{Back End:}} &\;\; \hat{V} =
\begin{cases}
\mathcal{G}_\theta\!\big(\mathcal{R}(P,\tau); I_t, L\big), &\!\! \text{generative}, \\
\mathcal{R}_{\text{3DGS}}(\tau), &\!\! \text{reconstructive},
\end{cases}
\end{aligned}
\label{eq:pipeline}
\end{equation}
where a coding agent $\mathcal{A}$ synthesizes $P$ from a read-only multimodal
evidence package $\mathcal{E}$, $\tau$ is a collision-free camera trajectory
planned toward the target in $L$, $\mathcal{R}$ renders an untextured structural
proxy video $V_{\text{proxy}} = \mathcal{R}(P,\tau)$ for a frozen generative
video renderer $\mathcal{G}_\theta$, and $\mathcal{R}_{\text{3DGS}}$ directly
renders novel views when a calibrated 3DGS scan is available. As illustrated in
Fig.~\ref{fig:method}, NavCWM separates navigation world modeling into a
front end for code-space scene synthesis and semantic path planning in
Secs.~\ref{sec:method-synthesis}--\ref{sec:method-planning} and a
back end for modular visual rendering in Sec.~\ref{sec:method-rendering}.
This training-free, causal front end runs either per episode from partial
history $I_{1:t}$ or once on a captured 3DGS scan, lifting $P$ once and
grounding each new instruction $L$ onto it.

\subsection{Grounded Code-Space Scene Synthesis}
\label{sec:method-synthesis}

\paragraph{Causal Multimodal Evidence}
Synthesizing 3D scenes from raw RGB alone readily induces scale drift and
layout errors. Strictly within the causal past interval $[1, t]$, we assemble a
read-only metric evidence package
$\mathcal{E} = (\mathcal{X}, \mathcal{C}, \mathcal{D}, \mathcal{S})$ comprising
metric-aligned 3D geometry $\mathcal{X}$, calibrated cameras
$\mathcal{C} = (C_i)_{i=1}^{t}$ with intrinsics and poses, and aligned depth
$\mathcal{D}$ extracted via the metric $\pi^3$ variant
Pi3X~\cite{wang2026pi3} per episode (or directly from calibrated multi-view
observations during one-time 3DGS scan lifting), alongside grounded
open-vocabulary 3D semantics $\mathcal{S}$ from SAM~3~\cite{carion2025sam3}.

\begin{figure}[t]
  \centering
  \includegraphics[width=0.97\columnwidth]{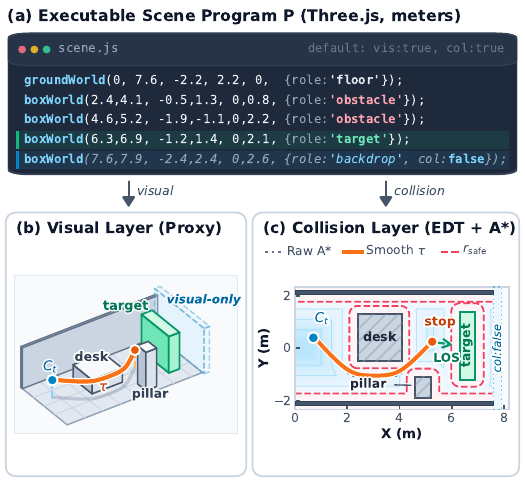}
  \vspace{-0.4em}
  \caption{\textbf{Executable 3D code space.}
  \textbf{(a)}~A metric Three.js scene program $P$ declares floors, obstacles, the target, and visual-only backdrops.
  \textbf{(b)}~The visual layer retains visual-only backdrops to render untextured white-box proxies preserving distant perspective.
  \textbf{(c)}~The collision layer builds a Euclidean distance transform (EDT) clearance field with margin $r_{\text{safe}}$ and plans a smooth A* trajectory $\tau$ to a line-of-sight stop pose.}
  \label{fig:code_space}
  \vspace{-0.45em}
\end{figure}

\paragraph{Executable Code-Space World State}
Rather than encoding layout in implicit latents, NavCWM represents the scene as
an executable Three.js scene program $P$ defining traversable floors,
obstacles, openings, targets, visual-only backdrops, and semantic landmarks from $\mathcal{S}$ in metric
coordinates. As shown in Fig.~\ref{fig:code_space}, visual and collision layers
are decoupled so non-collidable background structures preserve scene perspective
without restricting navigable space. Projecting $P$
under the calibrated cameras in $\mathcal{C}$ maintains pixel alignment with
$I_{1:t}$. Because $P$ is executable code, code-space clearance, visibility, and
reachability are evaluated deterministically while remaining inspectable across
both per-episode synthesis and lifted real-world scans.

\paragraph{Render-and-Compare with Deterministic Verification}
One-shot code generation from partial views rarely achieves tight geometric
alignment. Starting from an initial draft $P^{(0)} = \mathcal{A}(\mathcal{E}, L)$,
agent $\mathcal{A}$ refines $P$ over iterations $k = 0, 1, \dots$ via
render-and-compare:
\begin{equation}
\begin{aligned}
\tilde{I}^{(k)} &= \big(\mathcal{R}(P^{(k)}, C_i)\big)_{i=1}^{t}, \\
P^{(k+1)} &= \mathcal{A}\big(P^{(k)}, \mathcal{E}, \Delta(\tilde{I}^{(k)}, I_{1:t}, \mathcal{E}, L)\big),
\end{aligned}
\label{eq:render-compare}
\end{equation}
where $\mathcal{R}$ projects $P^{(k)}$ under $\mathcal{C}$ into rendered views
$\tilde{I}^{(k)}$, and comparator $\Delta$ evaluates multi-view geometric and
semantic alignment against $(I_{1:t}, \mathcal{E}, L)$. Before accepting $P$, a
programmatic verifier checks five deterministic criteria---\emph{target
grounding}, \emph{walkable floor hygiene}, \emph{3D point-cloud support}
(suppressing hallucinated geometry), \emph{in-loop A* reachability}, and
\emph{mesh/render-compare consistency}---returning structured diagnostics to
$\mathcal{A}$ for targeted parameter patching whenever a check fails.

\subsection{Collision-Free Code-Space Planning}
\label{sec:method-planning}

Over the explicit collision layer of $P$, NavCWM builds a role-separated 2D
occupancy grid~\cite{elfes1989occupancy} distinguishing walkable floors from
obstacles and computes an exact Euclidean distance transform (EDT) clearance
field inflated by safety margin $r_{\text{safe}}$. From the semantic layer, it
resolves a collision-free, line-of-sight stop pose at a semantic standoff from
the landmark specified by $L$, runs clearance-aware A*
search~\cite{hart1968astar} from the current pose of $C_t \in \mathcal{C}$, and
applies adaptive shortcutting and kinematic velocity profiling to produce a
smooth $H$-frame camera trajectory
$\tau = (T_s)_{s=1}^{H} \in \mathrm{SE}(3)^{H}$.

\begin{figure*}[t]
\centering
\includegraphics[width=0.89\textwidth]{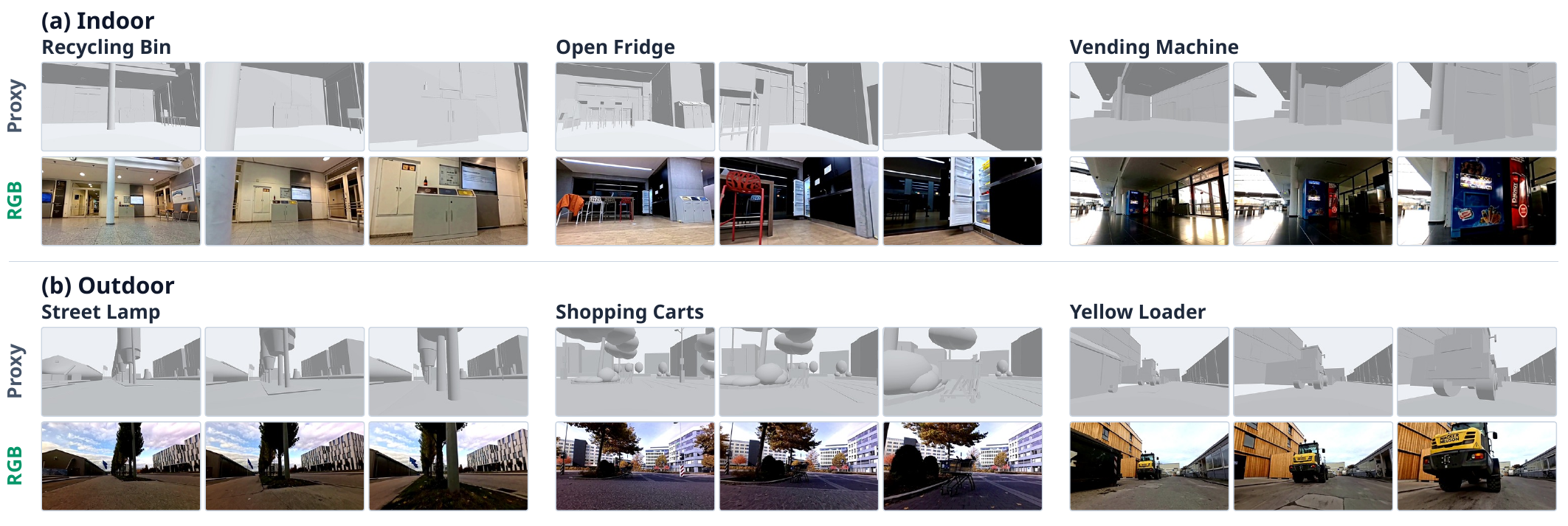}
\vspace{-0.6em}
\caption{\textbf{Proxy-to-RGB structural grounding across indoor and outdoor scenes.} Steered by untextured white-box proxies $V_{\text{proxy}}$ in the top \emph{Proxy} rows alongside $I_t$ and $L$, frozen MiniMax-H3 tracks planned turns, clears obstacles, and stops at semantic goals while preserving photorealistic appearance in the bottom \emph{RGB} rows $\hat{V}$ across six diverse \textbf{(a)}~indoor and \textbf{(b)}~outdoor scenes.}
\label{fig:qualitative}
\vspace{-0.5em}
\end{figure*}

\subsection{Decoupled Visual Rendering}
\label{sec:method-rendering}

\paragraph{Structural Proxy Rendering}
Rendering the visual layer of $P$ along $\tau$ under the calibrated intrinsics
from $\mathcal{C}$ deterministically produces a structural proxy video:
\begin{equation}
V_{\text{proxy}} = \mathcal{R}(P, \tau) \triangleq \big(\mathcal{R}(P, T_s)\big)_{s=1}^{H},
\label{eq:proxy}
\end{equation}
shaded as uniform untextured white-box geometry without synthetic colors that
could conflict with $I_t$. This proxy $V_{\text{proxy}}$ preserves multi-depth
3D parallax, occlusion boundaries, camera ego-motion, and task-relevant
landmarks along the planned route, supplying explicit metric motion cues that
text instructions alone cannot specify.

\paragraph{Early-Step Structural Conditioning}
While $V_{\text{proxy}}$ provides accurate 3D layout and motion, it lacks
photorealism; conversely, an unconstrained video generator often suffers from
motion collapse and spatial drift. To combine both strengths without retraining,
we condition a frozen reference-to-video diffusion renderer $\mathcal{G}_\theta$
on $V_{\text{proxy}}$, the current frame $I_t$ as an appearance anchor, and
instruction $L$ via early-step conditioning, following Code Video
Model~\cite{codevideomodel}. Starting from Gaussian latent noise
$z_0 \sim \mathcal{N}(\mathbf{0}, \mathbf{I})$, we inject proxy guidance only in
the first $K$ coarse denoising steps $\mathcal{G}_\theta^{(m)}$ for
$m = 1, \dots, M$ with $0 \le K \le M$, applying null structural conditioning
$\varnothing$ thereafter:
\begin{equation}
z_{m} =
\begin{cases}
\mathcal{G}_\theta^{(m)}\big(z_{m-1}; V_{\text{proxy}}, I_t, L\big), & m \le K, \\
\mathcal{G}_\theta^{(m)}\big(z_{m-1}; \varnothing, I_t, L\big), & m > K,
\end{cases}
\label{eq:early-step}
\end{equation}
and decode the predicted video via $\hat{V} = \mathrm{Dec}(z_M)$. This schedule
locks 3D layout and ego-motion early while late steps $m > K$ synthesize
textures and lighting from $I_t$ and $L$, cleanly separating roles: $L$
specifies semantics, $I_t$ anchors appearance, and $V_{\text{proxy}}$ enforces
metric motion.

\paragraph{Renderer-Agnostic Modularity}
Because $(P, \tau, V_{\text{proxy}})$ form a symbolic--geometric interface decoupled
from $\mathcal{G}_\theta$, the same front end transfers zero-shot across frozen video
renderers: general diffusion backbones use early-step conditioning via
Eq.~(\ref{eq:early-step}), whereas native proxy-conditioned renderers
(SolarRenderer~\cite{solarrenderer}, EdenLive~\cite{edenlive}) consume
$(V_{\text{proxy}}, I_t, L)$ directly. When a calibrated 3DGS
reconstruction~\cite{kerbl20233dgs} is available, $\tau$ directly drives
$\mathcal{R}_{\text{3DGS}}(\tau)$ using \texttt{gsplat}~\cite{ye2025gsplat}
without generating $V_{\text{proxy}}$.

\section{Experiments}
\label{sec:experiments}
\subsection{Setup}
\label{sec:exp-setup}
For video world modeling, we evaluate on all 125 Target-Bench test
segments~\cite{targetbench} under the official VGGT-Omega metric-trajectory
protocol, reporting Weighted Overall (WO$\uparrow$),
Average/Final Displacement Error (ADE/FDE, m$\downarrow$), Miss Rate
(MR, \%$\downarrow$), Soft Endpoint (SE$\uparrow$), and Approach Consistency
(AC$\uparrow$). From each first-to-current prefix $I_{1:t}$ with Gemini~4 Neon
as coding agent $\mathcal{A}$ (fixed random seed), our front end drives three
frozen video renderers: reference-to-video MiniMax-H3 (R2V, $M{=}20$ steps,
default $K{=}5$), alongside native proxy-conditioned
SolarRenderer~\cite{solarrenderer} and EdenLive~\cite{edenlive}.

\subsection{Main Result: Code Proxy vs.\ No-Proxy Conditioning}
\begin{table}[!t]
\centering
\caption{Quantitative Comparison on Target-Bench ($n{=}125$). ZS denotes zero-shot evaluation. Results marked with $\dagger$ are from Target-Bench~\cite{targetbench}; best zero-shot results are in bold.}
\label{tab:main}
\setlength{\tabcolsep}{1.8pt}
\renewcommand{\arraystretch}{0.91}
\footnotesize
\begin{tabular}{@{}clcccccc@{}}
\toprule
ZS & Method & WO$\uparrow$ & ADE$\downarrow$ & FDE$\downarrow$ & MR$\downarrow$ & SE$\uparrow$ & AC$\uparrow$ \\
\midrule
\multicolumn{8}{@{}l@{}}{\emph{Target-Bench fine-tuned reference:}} \\
$\times$ & \emph{Wan2.2-5B}$^\dagger$ & \emph{0.478} & \emph{0.400} & \emph{0.915} & \emph{9.0} & \emph{0.393} & \emph{0.834} \\
\midrule
\multicolumn{8}{@{}l@{}}{\emph{General video world models:}} \\
$\checkmark$ & Sora 2$^\dagger$ & 0.248 & 0.984 & 1.787 & 40.0 & 0.173 & 0.835 \\
$\checkmark$ & Veo 3.1$^\dagger$ & 0.225 & 1.981 & 3.055 & 53.3 & 0.181 & 0.425 \\
$\checkmark$ & Wan2.2-Flash$^\dagger$ & 0.385 & 0.718 & 1.297 & 22.3 & 0.309 & 0.767 \\
$\checkmark$ & MiniMax-H3 & 0.277 & 0.959 & 1.650 & 35.2 & 0.204 & 0.576 \\
\midrule
\multicolumn{8}{@{}l@{}}{\emph{Navigation video world models (language-guided):}} \\
$\checkmark$ & SparseVideoNav & 0.197 & 1.282 & 2.363 & 45.0 & 0.121 & 0.621 \\
$\checkmark$ & ImagiNav & 0.150 & 1.153 & 2.185 & 51.8 & 0.066 & \textbf{0.988} \\
\midrule
\multicolumn{8}{@{}l@{}}{\textbf{NavCWM (Ours):}} \\
$\checkmark$ & \hspace*{0.8em}w/ MiniMax-H3 & \textbf{0.503} & 0.530 & 0.820 & \textbf{11.3} & \textbf{0.430} & 0.903 \\
$\checkmark$ & \hspace*{0.8em}w/ SolarRenderer & 0.496 & \textbf{0.503} & 0.848 & \textbf{11.3} & 0.424 & 0.851 \\
$\checkmark$ & \hspace*{0.8em}w/ EdenLive & 0.494 & 0.559 & \textbf{0.818} & 12.2 & 0.424 & 0.849 \\
\bottomrule
\end{tabular}
\vspace{-0.4em}
\end{table}

In Table~\ref{tab:main}, code-proxy conditioning lifts frozen MiniMax-H3 from
0.277 to 0.503 WO without weight updates, cutting MR by over two-thirds from
35.2\% to 11.3\% and more than doubling SE from 0.204 to 0.430. NavCWM improves
over unguided MiniMax-H3 on every metric, leads all zero-shot baselines, and
surpasses fine-tuned Wan2.2-5B on WO, FDE, SE, and AC (0.503 vs.\ 0.478 WO;
Wan2.2-5B retains lower ADE and MR).

\begin{figure*}[t]
\centering
\includegraphics[width=0.90\textwidth]{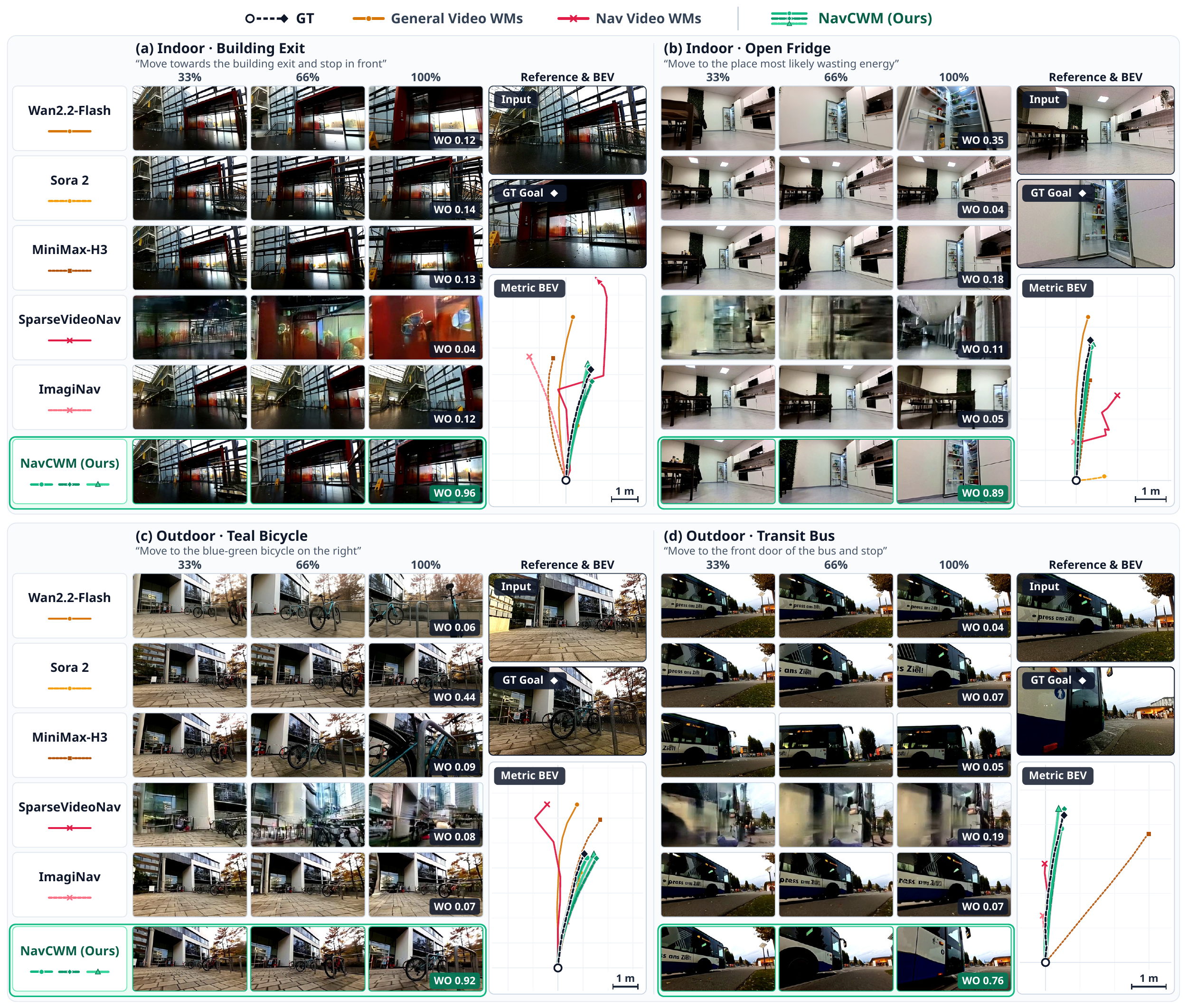}
\vspace{-0.6em}
\caption{\textbf{Qualitative comparison on Target-Bench:} \textbf{(a--b)}~indoor, \textbf{(c--d)}~outdoor. Per panel, \emph{left:} unguided baselines often stop short, overshoot, or drift, whereas NavCWM \emph{w/ MiniMax-H3} lifts unguided MiniMax-H3 WO$\uparrow$ from 0.05--0.18 to 0.76--0.96; \emph{right:} reference views and metric bird's-eye-view (BEV) trajectories comparing ground truth (GT) against emerald curves for solid $\bullet$ \emph{MiniMax-H3}, dashed $\blacklozenge$ \emph{SolarRenderer}, and dash-dotted $\triangle$ \emph{EdenLive}.}
\label{fig:baseline_cmp}
\vspace{-0.5em}
\end{figure*}

Qualitatively, Fig.~\ref{fig:qualitative} illustrates proxy-to-RGB grounding
across six indoor and outdoor Target-Bench scenes: from untextured white-box
geometry alone, frozen MiniMax-H3 executes planned turns, clears pillars and
furniture, and stops at the planned distance while retaining lighting,
reflections, and materials from $I_t$. Figure~\ref{fig:baseline_cmp} contrasts
NavCWM against five zero-shot baselines across indoor and outdoor scenes:
unguided models often stop short, overshoot obstacles, or drift off course,
whereas proxy-guided MiniMax-H3 tracks ground truth (GT) cleanly and converges
near each goal in the bird's-eye-view (BEV) plots.

\subsection{Ablation: Proxy Injection and Proxy Content}
\begin{figure*}[t]
\centering
\includegraphics[width=0.86\textwidth]{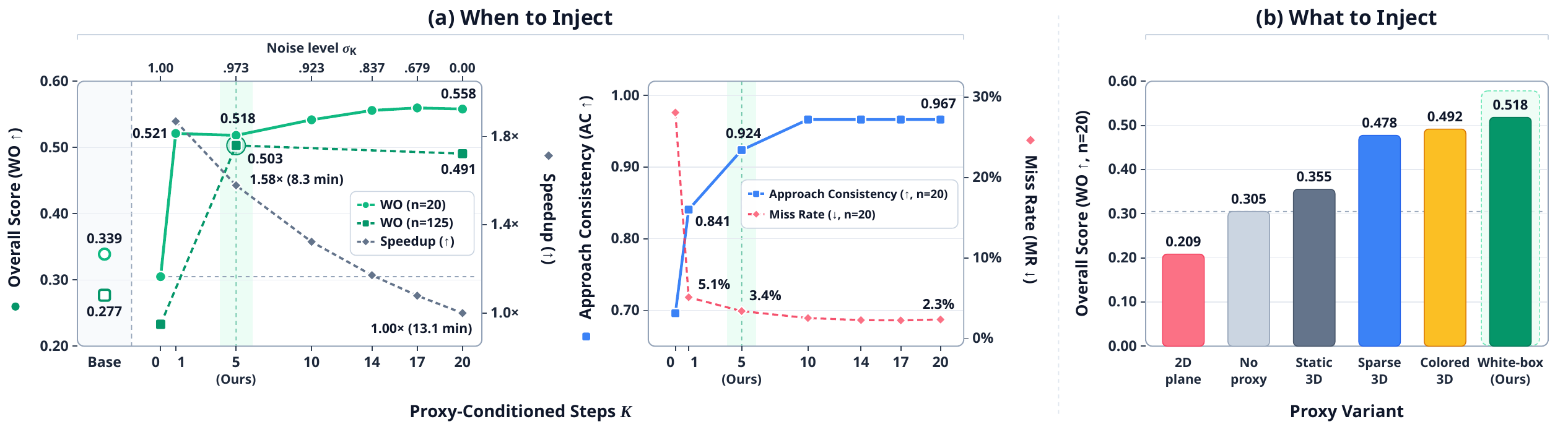}
\vspace{-0.75em}
\caption{\textbf{Ablations on injection schedule and proxy content} using frozen MiniMax-H3 over 20 steps, with top axis showing noise level $\sigma_K$. \textbf{(a)}~\emph{When to Inject:} $n{=}20$ sweep with $n{=}125$ markers at Base and $K\in\{0,5,20\}$; $K{=}5$ sharply reduces single-step AC/MR instability and outperforms $K{=}20$ on full $n{=}125$ at 1.58$\times$ speedup. \textbf{(b)}~\emph{What to Inject:} On $n{=}20$ at $K{=}5$, neutral \emph{White-box (Ours)} (0.518 vs.\ \emph{No proxy}, 0.305 dashed line) leads metric accuracy via complete, appearance-free 3D parallax.}
\label{fig:ablation}
\vspace{-0.6em}
\end{figure*}
In Table~\ref{tab:ablation}a and Fig.~\ref{fig:ablation}a, disabling proxy
guidance under null-proxy R2V ($K{=}0$) drops full-test WO to 0.233---below
native first-frame MiniMax-H3 (0.277)---confirming that the gains stem from
structural conditioning rather than R2V scaffolding or prompting. Across the
$n{=}20$ sweep, even a single conditioned step establishes coarse heading, yet
execution remains erratic until early structure is locked over several steps;
moderate early injection at $K{=}5$ stabilizes trajectory adherence while
leaving late denoising free to synthesize photorealistic appearance. On all 125
scenes, $K{=}5$ reaches 0.503 WO and outperforms full-schedule conditioning at
$K{=}20$ across every metric while running 1.58$\times$ faster, showing that
late proxy tokens over-constrain fine textures and harm metric decoding rather
than improving control.

\begin{table}[!t]
\centering
\caption{Ablation Study on Injection Schedule and Proxy Content. All variants use frozen MiniMax-H3 over 20 steps; best results per block are in bold.}
\label{tab:ablation}
\setlength{\tabcolsep}{2.0pt}
\renewcommand{\arraystretch}{0.91}
\footnotesize
\begin{tabular}{@{}lcccccc@{}}
\toprule
Variant & WO$\uparrow$ & ADE$\downarrow$ & FDE$\downarrow$ & MR$\downarrow$ & SE$\uparrow$ & AC$\uparrow$ \\
\midrule
\multicolumn{7}{@{}l@{}}{\emph{(a)~Injection schedule (full test set, $n{=}125$):}} \\
MiniMax-H3 base & 0.277 & 0.959 & 1.650 & 35.2 & 0.204 & 0.576 \\
NavCWM w/o proxy ($K{=}0$) & 0.233 & 1.274 & 2.413 & 41.6 & 0.166 & 0.662 \\
\textbf{NavCWM ($K{=}5$, Ours)} & \textbf{0.503} & \textbf{0.530} & \textbf{0.820} & \textbf{11.3} & \textbf{0.430} & \textbf{0.903} \\
NavCWM ($K{=}20$, full) & 0.491 & 0.547 & 0.870 & 12.0 & 0.417 & 0.893 \\
\midrule
\multicolumn{7}{@{}l@{}}{\emph{(b)~Proxy content (stratified subset, $n{=}20$, $K{=}5$):}} \\
2D plane & 0.209 & 1.531 & 2.203 & 58.1 & 0.166 & \textbf{0.957} \\
No proxy ($K{=}0$) & 0.305 & 0.730 & 1.324 & 28.1 & 0.216 & 0.696 \\
Static 3D & 0.355 & 0.948 & 1.467 & 35.2 & 0.312 & 0.706 \\
Sparse 3D & 0.478 & 0.555 & 0.946 & 10.4 & 0.399 & \textbf{0.957} \\
Colored 3D & 0.492 & 0.547 & 0.935 & 10.9 & 0.411 & 0.905 \\
\textbf{Full white-box (Ours)} & \textbf{0.518} & \textbf{0.438} & \textbf{0.694} & \textbf{3.4} & \textbf{0.436} & 0.924 \\
\bottomrule
\end{tabular}
\end{table}

\begin{figure}[!t]
\centering
\includegraphics[width=\columnwidth]{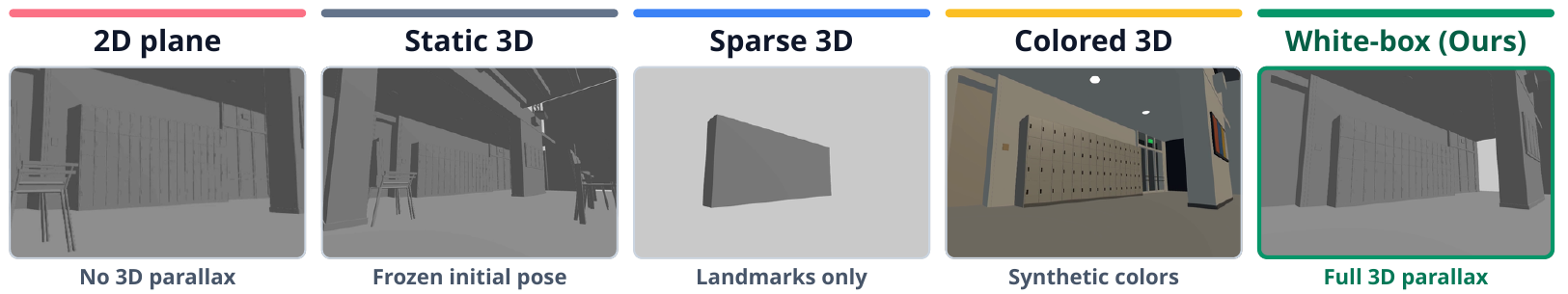}
\vspace{-0.9em}
\caption{\textbf{Proxy variants in Fig.~\ref{fig:ablation}b} (Target-Bench \#348, mid-route). Neutral \emph{White-box (Ours)} supplies full 3D parallax without synthetic colors.}
\label{fig:proxy_variants}
\vspace{-0.6em}
\end{figure}

In Table~\ref{tab:ablation}b and Figs.~\ref{fig:ablation}b and~\ref{fig:proxy_variants}, controlled ablations
under fixed route, camera, and seed show that \emph{Full white-box} leads five
of six metrics at 0.518 WO and 3.4\% MR, isolating four distinct
effects. \emph{2D plane}, a planar homography warp of the initial white-box
frame lacking depth disparity, performs worse than \emph{No proxy} at 0.209 WO
and induces severe early stalling, which artificially inflates AC without
advancing toward the goal. \emph{Static 3D} freezes initial-view geometry
without ego-motion, providing weak spatial anchoring yet worsening endpoint
misses. \emph{Sparse 3D} retains only nearby landmarks along the route without
room or floor shells and recovers partial parallax at 0.478 WO, whereas complete
enclosure and ground geometry cuts terminal error and reduces misses threefold.
Finally, \emph{Colored 3D} introduces synthetic colors and lighting that clash
with $I_t$, degrading control and more than tripling misses. Together, these
results reveal that frozen video renderers decode appearance-free, multi-depth
3D parallax across complete scene structure rather than copying proxy pixels or
raw 2D flow.

\subsection{Separability of State and Generation}
\label{sec:exp-transfer}
Because spatial state resides in executable 3D code rather than generator
weights, identical $(P, \tau, V_{\text{proxy}})$ transfer zero-shot across
frozen renderers without retraining or proxy modification. In
Table~\ref{tab:main}, pairing this shared front end with frozen MiniMax-H3 and
native proxy renderers SolarRenderer and EdenLive forms a tight 0.494--0.503 WO cluster whose 0.009
spread sits well within the 0.024 per-episode standard error, with each renderer
leading at least one zero-shot metric. Across both indoor and outdoor episodes,
all three renderers trace nearly coinciding BEV trajectories along each
planned path and converge near the GT goal (Fig.~\ref{fig:baseline_cmp}, BEV
plots), confirming that trajectory geometry is governed by the code front end
rather than visual synthesis.

\subsection{Comparison with Navigation Video World Models}
Turning to specialized navigation world models in Table~\ref{tab:main},
action-conditioned NWM~\cite{nwm} takes no language input, so we evaluate the
language-guided SparseVideoNav~\cite{sparsevideonav} and
ImagiNav~\cite{imaginav} in this zero-shot cross-domain setting. Outside their
training domains, both preserve heading better than metric scale:
ImagiNav attains the highest zero-shot AC at 0.988 through conservative steps
along the initial heading, though its metric displacement remains limited,
whereas SparseVideoNav maintains reasonable heading alignment while tending to
overshoot or drift under domain shift. These observations suggest that coupling
visual appearance and motion in pixel or latent space makes metric scale
sensitive to domain shift, whereas explicit 3D code proxies help calibrate both
heading and travel distance prior to rendering.

\subsection{Executable Spatial Memory for Real-World VLN}
\label{sec:exp-data-engine}
\begin{figure*}[t]
\centering
\includegraphics[width=0.89\textwidth]{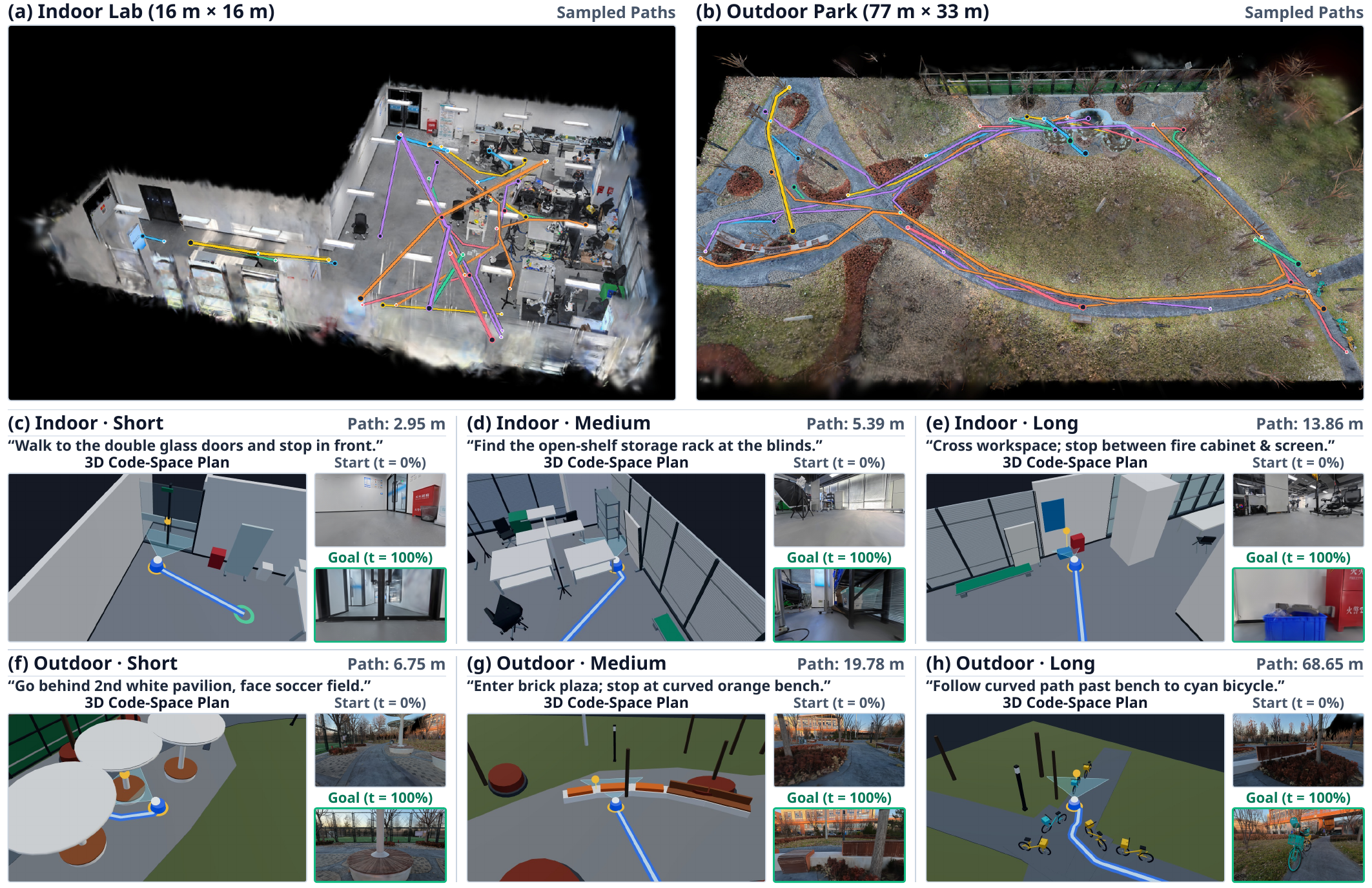}
\vspace{-0.5em}
\caption{\textbf{Executable spatial memory and VLN data engine in real scans.} Lifting each captured scan once into executable 3D scene code enables batch collision-free planning in code space from open-vocabulary instructions across \textbf{(a)}~Indoor Lab and \textbf{(b)}~Outdoor Park (sampled routes). \textbf{(c--h)}~Short-to-long code-space stop poses (2.95--68.65\,m; meshes colored for clarity) align closely with 3DGS-rendered start/goal views.}
\label{fig:data_engine}
\vspace{-0.45em}
\end{figure*}
\begin{table}[!t]
\centering
\caption{Spatial Memory for Navigation in Real Scans ($n{=}100$/scene). Nav.\ error NE (m); 1\,m SR/SPL, raw point-cloud collision-free CF, and joint FullSR (\%); Toks: mean prompt tokens.}
\label{tab:data_engine}
\setlength{\tabcolsep}{2.2pt}
\renewcommand{\arraystretch}{0.91}
\footnotesize
\begin{tabular}{@{}lcccccc@{}}
\toprule
Method & NE$\downarrow$ & SR$\uparrow$ & SPL$\uparrow$ & CF$\uparrow$ & FullSR$\uparrow$ & Toks$\downarrow$ \\
\midrule
\multicolumn{7}{@{}l@{}}{\emph{(a)~Indoor Lab (16${\times}$16\,m, $n{=}100$):}} \\
ReMEmbR~\cite{anwar2025remembr} & 2.65 & 47 & 42.8 & \textbf{100} & 47 & 65k \\
\textbf{NavCWM (Ours)} & \textbf{0.81} & \textbf{75} & \textbf{71.6} & 85 & \textbf{64} & \textbf{26k} \\
\midrule
\multicolumn{7}{@{}l@{}}{\emph{(b)~Outdoor Park (77${\times}$33\,m, $n{=}100$):}} \\
ReMEmbR~\cite{anwar2025remembr} & 5.00 & 11 & 10.6 & \textbf{94} & 11 & 46k \\
\textbf{NavCWM (Ours)} & \textbf{2.13} & \textbf{53} & \textbf{48.5} & 75 & \textbf{46} & \textbf{14k} \\
\bottomrule
\end{tabular}
\vspace{-0.4em}
\end{table}
Beyond Target-Bench, we evaluate NavCWM on real-world 3DGS
scans~\cite{kerbl20233dgs} of a 16${\times}$16\,m indoor lab and a
77${\times}$33\,m outdoor park across 100 open-vocabulary instructions each.
Ground-truth stop poses are manually annotated on the raw scans and verified in
3DGS views, while obstacle fields are extracted directly from the raw dense
point clouds, independent of the lifted scene code. Lifting each scan once into
executable 3D code enables long-horizon planning and novel-view synthesis in
Fig.~\ref{fig:data_engine}, reversing trajectory-first
captioning~\cite{yu2026onescan}. Under identical 3DGS inputs and the same
vision-language model backbone, we compare against ReMEmbR~\cite{anwar2025remembr}, which queries
tour keyframes over a point-cloud occupancy grid across multiple turns, whereas
NavCWM grounds directly in code in one turn.

In Table~\ref{tab:data_engine}, episodic retrieval struggles to disambiguate
similar landmarks or resolve exact standoffs over long horizons, dropping to
0\% strict 1\,m success on long outdoor routes where NavCWM retains 44\%.
Explicit 3D code cuts mean indoor error from 2.65\,m to 0.81\,m and lifts
outdoor success nearly fivefold from 11\% to 53\%. Geometric safety reflects an
expected abstraction trade-off: planning directly on the raw point-cloud grid
gives ReMEmbR higher raw clearance, because compact code primitives occasionally
omit small obstacles and overhanging shrubs. Even so, stronger goal grounding
keeps NavCWM ahead on joint safe arrival at 64\% indoors and 46\% outdoors, and
pairing code-grounded goals with point-cloud planning could combine both
strengths. Encoding global layout and semantics in one program also uses
2.5--3.3$\times$ fewer prompt tokens (14k in the sparser park vs.\ 26k indoors)
and lets batch A* plan 100 routes per scene in under 5\,s.

\section{Discussion}
\label{sec:discussion}
\noindent\textbf{1) Factorizing geometry from appearance.}
Video world models face metric drift when pixel rollouts entangle appearance
with 3D layout. Untextured proxies show strong visual priors already reside in
frozen generators; explicit metric structure lets perception and visual
foresight share one code, echoing human ventral temporal
cortex~\cite{wadia2026}.

\vspace{2pt}
\noindent\textbf{2) Deterministic state, replaceable renderers.}
Navigation demands deterministic clearance, whereas visual generators
prioritize appearance and evolve rapidly. Keeping spatial state in executable
code rather than weights sets a clean boundary: rendering artifacts cannot
corrupt deterministic code-space collision checks, while frozen video models
and 3DGS share the same verified layout.

\vspace{2pt}
\noindent\textbf{3) From passive memory to executable worlds.}
Black-box video rollouts, raw 3D scans, and episodic keyframe memories do not
directly expose a compact, verifiable spatial state for long-horizon planning.
Lifting observations once into executable 3D code turns passive records into a
queryable spatial memory that anchors open-vocabulary instructions in metric
geometry and complements trajectory-first pipelines~\cite{yu2026onescan}.
Because global layout and named landmarks reside in one concise program,
queries resolve in a single turn and plan deterministically without retrieval
drift.

\section{Conclusion}
\label{sec:conclusion}
This paper introduced NavCWM, a training-free navigation world model that
decouples semantic path planning from visual rendering through executable 3D
scene code. Extensive experiments and controlled ablations reveal that coupled
pixel rollouts struggle to preserve metric scale across domains, whereas
complete untextured 3D geometry reliably steers frozen video renderers without
retraining. By keeping explicit spatial state in executable code, NavCWM
isolates deterministic code-space collision checks and planning from appearance
synthesis and transfers cleanly across generative and reconstructive back ends,
achieving strong training-free performance on Target-Bench. In captured
real-world scans, the same code representation serves as an executable spatial
memory that outperforms episodic visual memory in long-horizon navigation
accuracy and query efficiency while enabling instruction-driven 3DGS data
generation. We believe this code-space paradigm offers an inspectable, renderer-agnostic
foundation for generalizable embodied world models.

\clearpage
\bibliographystyle{IEEEtran}
\bibliography{main}

\end{document}